\documentclass[runningheads]{llncs}
\usepackage{graphicx}
\usepackage{multirow}
\usepackage[T1]{fontenc}
\usepackage[super]{nth}

\usepackage{dirtytalk}

\usepackage{amssymb}
\usepackage{mathtools}
\usepackage{dsfont}

\newcommand{\R}[0]{\mathds{R}} % real numbers
\newcommand{\C}[0]{\mathds{C}} % complex numbers
\providecommand{\mb}[1]{\mathbf{#1}}
\providecommand{\mbx}{\mb{x}}
\providecommand{\mby}{\mb{y}}

\begin{document}
\title{Universal Undersampled MRI Reconstruction}
\titlerunning{Universal Undersampled MRI Reconstruction}

\author{Xinwen Liu\inst{1} \and
Jing Wang\inst{2} \and Feng Liu\inst{1} \and S.Kevin Zhou\inst{3}}
\authorrunning{X.Liu, et al.}
\institute{School of Information Technology and Electrical Engineering, \\The University of Queensland, Australia.\\
\and CSIRO, Australia. \\
\and Institute of Computing Technology, Chinese Academy of Sciences, China.}

\maketitle
\begin{abstract}
Deep neural networks have been extensively studied for undersampled MRI reconstruction. While achieving state-of-the-art performance, they are trained and deployed specifically for one anatomy with limited generalization ability to another anatomy. Rather than building multiple models, a universal model that reconstructs images across different anatomies is highly desirable for efficient deployment and better generalization. Simply mixing images from multiple anatomies for training a single network does not lead to an ideal universal model due to the statistical shift among datasets of various anatomies, the need to retrain from scratch on all datasets with the addition of a new dataset, and the difficulty in dealing with imbalanced sampling when the new dataset is further of a smaller size. In this paper, for the first time, we propose a framework to \textbf{learn a universal deep neural network for undersampled MRI reconstruction}. Specifically, anatomy-specific instance normalization is proposed to compensate for statistical shift and allow easy generalization to new datasets. Moreover, the universal model is trained by distilling knowledge from available independent models to further exploit representations across anatomies. Experimental results show the proposed universal model can reconstruct both brain and knee images with high image quality. Also, it is easy to adapt the trained model to new datasets of smaller size, i.e., abdomen, cardiac and prostate, with little effort and superior performance.
\end{abstract}
\section{Introduction}
Magnetic Resonance Imaging (MRI) is a non-invasive imaging modality with superior soft-tissue visualization, but it is slow due to the sequential acquisition scheme. One direct approach to accelerating MRI is to sample less k-space data; however, the images directly reconstructed from the undersampled data are blurry and full of artifacts. Conventionally, compressed sensing (CS) \cite{lustig2007sparse} has been exploited to recover high-quality images from the undersampled measurements. Recently, deep learning (DL) \cite{aggarwal2018modl,hammernik2018learning,knoll2019assessment,qin2018convolutional,schlemper2017deep,sriram2020end,sriram2020grappanet,wang2016accelerating,yang2017dagan,yang2016deep,zhou2019handbook} has been extensively studied for higher accuracy and faster speed.

While DL achieves state-of-the-art reconstruction performance \cite{knoll2020advancing,recht2020using}, such success is often within the same dataset; in other words, the network trained on a dataset specific to one anatomy hardly generalize to datasets of other anatomies. To reconstruct images of various anatomies, the existing DL-based reconstruction framework usually needs to train an individual model for each dataset separately; and then multiple models are deployed on machines for application \cite{sriram2020end}. There are drawbacks to this approach. First, numerous models lead to a large number of parameters, which makes it hard to deploy on commercial MRI machines.Second, datasets of various anatomies contain shared prior knowledge \cite{lustig2007sparse}; separately trained networks do not exploit such common knowledge, which may limit the effectiveness of the trained models. While authors in \cite{ouyang2019generalising} attempted to generalize DL-based reconstruction, the network is trained purely on natural images with limited anatomical knowledge. Therefore, it is highly desirable to train a universal model that exploits cross-anatomy common knowledge and conducts reconstruction for various datasets, even when some datasets are very small.

A straightforward way to have one model for all anatomies is by simply mixing various datasets during training. While this reduces the number of parameters, it unfortunately faces a variety of challenges. As demonstrated in \cite{huang20193d,zhou2020review}, various datasets have different distributions, that is, there exist statistical shifts among datasets. Training a network with datasets of varying distributions degrades performance; therefore, simply mixing all data for training does not lead to an ideal universal model. In addition, when a dataset of new anatomy becomes available later, the model needs to be retrained from scratch on all datasets. It is time-consuming and storage-hungering for training and storing all the datasets. Finally, while there exist large-scale datasets, the newly collected datasets are normally small-scale. Mixing datasets of varying sizes for training causes imbalanced sampling  \cite{zhou2020review}, so the network may not be sufficiently optimized and the performance is questionable.

To address the issues mentioned above, for the first time, we propose a framework to train a \textbf{universal network}, which can reconstruct images of different anatomies and is generalizable to new anatomies easily. The base of the universal model is a convolutional network regularly used for MRI reconstruction, capturing commonly shared knowledge. To compensate for statistical shift and capture the knowledge specific to each dataset, we design an \textbf{Anatomy-SPecific Instance Normalization (ASPIN)} module to accompany the base model for reconstructing multiple anatomies. When a new anatomy dataset is available, a new set of ASPIN parameters is inserted and trained with the base of universal model fixed. The new ASPIN only needs few parameters and can be trained fast. Moreover, we propose to distil information from available models, trained purely on one anatomy, to the universal model. \textbf{Model distillation} allows the universal network to absorb multi-anatomy knowledge. In this study, we used public large-scale brain and knee data as currently available datasets. We took the abdomen, cardiac and prostate datasets as the new-coming small datasets. The experimental results show that the universal model can recover brain and knee images with better performance and fewer parameters than separately trained models.  When generalizing to the new cardiac dataset, and much smaller abdomen and prostate datasets, the universal model also presents superior image qualities to the individually trained models.
\section{Methods}
\subsection{The Overall Framework}
Reconstructing an image $\mbx \in \C^N$ from undersampled measurement  $\mby \in \C^M$ ($M \ll N$) is an inverse problem, which is formulated as \cite{schlemper2017deep,zhou2019handbook}: 

\begin{equation}
  \label{sparse_coding}
\begin{aligned}
& \underset{\mbx}{\text{min}}
& & \mathcal{R}(\mbx) + \lambda \| \mby - \mb{F}_u \mbx \|^2_2,
\end{aligned}
\end{equation}

\noindent where $\mathcal{R}$ is the regularisation term on $\mbx$, $\lambda \in \R$ denotes the balancing coefficient between $\mathcal{R}$ and data consistency (DC) term, $ \mb{F}_u \in \C^{M\times N}$ is an undersampled Fourier encoding matrix. In model-based learning, Eq.~\eqref{sparse_coding} is incorporated in the network architecture with $\mathcal{R}$ approximated by the convolutional layers. Taking D5C5 in reference \cite{schlemper2017deep} as an example, Eq. \eqref{sparse_coding} is unrolled into a feed-forward network path, cascading five inter-leaved DC blocks and learnable convolutional blocks that capture the prior knowledge of the training data. 

Let ${\{d_1, d_2, ..., d_A\}}$ be $A$ datasets of different anatomies to be reconstructed. Based $d_a$ for an anatomy $a$, a convolutional network $f_a$ is trained to recover its image. For example, in D5C5, $f_a$ comprises \{CNN-t; t$=$1:5\}, where CNN-t is a learnable convolutional block. Conventionally, to recovery images of $A$ anatomies, different networks $\{f_1, f_2,..., f_A\}$ are trained, one for each anatomy. In this paper, we aim to train a universal network $f_u$ that simultaneously recovers the images of various anatomies, with each convolutional block captures both common and anatomy-specific knowledge. We name such a block as a universal CNN (uCNN); overall $f_u$ is composed of \{uCNN-t; t$=$1:5\}. 

The universal network $f_u$ in this work is designed to achieve two aims: (A1) it can recover images of different anatomies exemplified by the currently available datasets, which are brain and knee datasets in this study; (A2) when the dataset of new anatomy is available, the trained $f_u$ can be generalized to the new dataset by adding a very few number of parameters.

\begin{figure}[t]
\includegraphics[width=\textwidth]{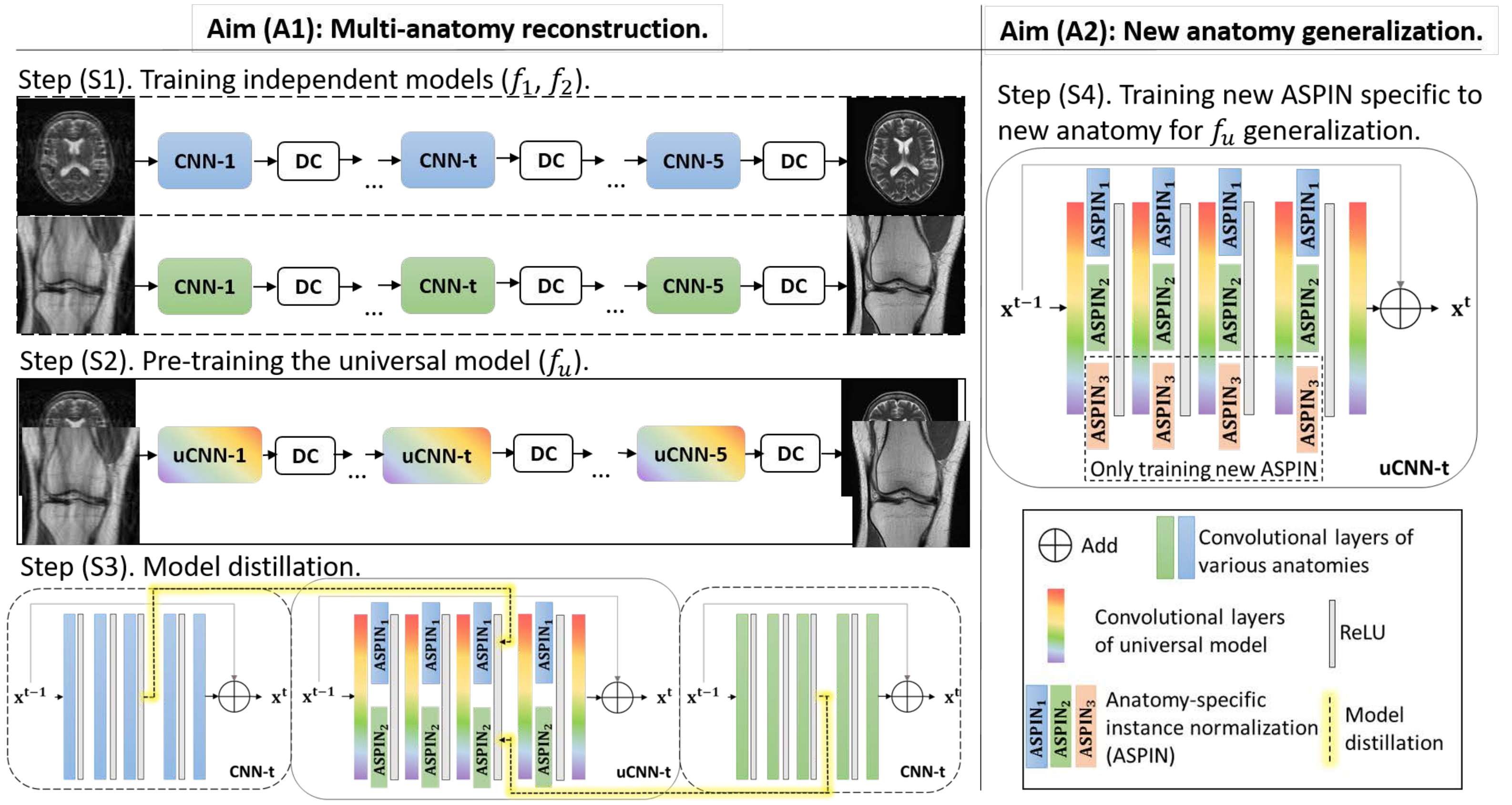}
\caption{The overall framework for universal undersampled MRI reconstruction.} \label{fig1}
\end{figure}

Fig. \Ref{fig1} depicts the overall framework of training $f_u$, which is comprised of four steps. (S1) We train two D5C5 networks ($f_1, f_2$) separately for brain and knee image recovery, and name them as independent models. These models are trained following \cite{schlemper2017deep}, where CNN-t contains five convolutional layers with four ReLU layers and the residual connection. (S2) We pre-train $f_u$ with both datasets. As shown in Fig. \Ref{fig1} Step (S2), CNN-t is replaced with uCNN-t, where two ASPIN modules are inserted, one for each anatomy. Although multi-anatomy knowledge is captured in this step, representation of single anatomy has not yet been fully utilized. (S3) We further optimize the pre-trained $f_u$ with model distillation. As illustrated in Fig. \Ref{fig1} Step (S3), the knowledge captured in the third layer of CNN-t in $f_1$ and $f_2$ is transferred to that of uCNN-t. After this step, the universal model can perform reconstruction on both datasets. (S4) When the model is needed for a new anatomy, we insert additional ASPIN to adapt the trained $f_u$ to the new dataset. The new dataset is only used to optimize the parameters of the newly inserted ASPIN, whereas all the other parameters are fixed. The generalized universal model learn to reconstruct new anatomy without forgetting the knowledge of old anatomies, capable of multi-anatomy reconstruction. Overall, Steps (S1-S3) achieves aim (A1), and Step (S4) achieves aim (A2). 

\subsection{Anatomy-SPecific Instance Normalization (ASPIN)}
One issue of mixing data from all anatomies as the training dataset is the statistical shift among datasets of various anatomies. To compensate for the distribution shift, we design the ASPIN module in the network.

Instance normalization (IN), proposed in \cite{huang2017arbitrary,ulyanov2016instance}, uses the statistics of an instance to normalize features. 
Let the output activation of a given layer be $h \in \mathbb{R}^{C\times H\times W}$, where $C$ is the number of channels, $H$ is the height of the feature map, and $W$ is the width of the feature map. Here we ignore the batch dimension for simplicity. The IN function is defined as:

\begin{equation}
\textrm{IN}(h)= \gamma\left(\frac{h-\mu(h)}{\sigma(h)}\right)+\beta,
\end{equation}

\noindent where $\gamma$ and $\beta$ are the learnable affine parameters, $\mu(h)$ and $\sigma(h)$ are the mean and standard deviation of the feature map computed across spatial dimensions ($H\times W$) independently for each channel.

ASPIN is designed by stacking multiple parallel IN functions, each of which is reserved for one anatomy. Specifically, for anatomy $d_a$, anatomy-specific affine parameters $\gamma_a$ and $\beta_a$ are learnt during the training process. The specific ASPIN function for anatomy $a$ is defined as: 
\begin{equation}
\textrm{ASPIN}_a(h)= \gamma_a\left(\frac{h-\mu(h)}{\sigma(h)}\right)+\beta_a,~a = 1,2,...,A.
\end{equation}

\noindent where $a$ is the index of anatomy corresponding to the current input image. It is expected that ASPIN captures anatomy-specific knowledge by learning the separate affine parameters for each anatomy. ASPIN allows the antecedent convolutional layer to capture anatomy-invariant knowledge, and the anatomy-specific information is presented by exploiting the captured statistics for the given anatomy. In this way, ASPIN is designed to compensate for the statistical difference of dataset distributions from various anatomies; and the sharable knowledge across anatomies is preserved with the shared convolutional layers.

\subsection{Model Distillation}
Independent models trained purely on a single anatomy capture more precise representation specifically for one dataset. Transferring the information from the fully-trained independent models to the universal model enables the network to capture more knowledge for each anatomy. To exploit the independent models' representations, we propose model distillation to further optimize the universal model.

The trained $f_1$ and $f_2$ in Step (S1) capture the knowledge of one anatomy. As in Fig. \Ref{fig1} Step (S3), we distil the knowledge learnt in the third layer of each CNN-t in $f_1$ and $f_2$ to that of each uCNN-t in the pre-trained $f_u$. The distillation is conducted by minimizing the attention transfer loss \cite{murugesan2020kd} between the independent models and the universal model. Let $m_{c} \in \mathbb{R}^{ H\times W}, c = 1,2,...,C$ be the $c^{th}$ channel of the output activation $h$. The spatial attention map of $h$ is defined as $O = \sum_{c=1}^{C}|m_{c}|^{2}$. In each cascade $t$, the attention transfer loss is expressed as:

\begin{equation}
  \label{att_loss}
    l_{AT}^{t} = ||\frac{O_{I}^{t}}{||O_{I}^{t}||_{2}} - \frac{O_{U}^{t}}{||O_{U}^{t}||_{2}}||_{1},
\end{equation}

\noindent where $O_{I}^{t}$ and $O_{U}^{t}$ represent the spatial attention map of the third layer in CNN-t in independent models ($f_1$ or $f_2$) and in uCNN-t in $f_u$, respectively. Overall, the attention transfer loss for D5C5 is $L_{AT} = \sum_{t=1}^{5} l_{AT}^{t}$. 

\subsection{Network Training Pipeline}
Individual models in Step (S1) are trained with mean-absolute-error (MAE) loss ($L_{MAE}$) on each of the datasets separately, to fully learn anatomy-specific knowledge. In Step (S2), the universal model is pre-trained on both datasets with $L_{MAE}$. During the training, we sample a batch of data from the same anatomy in a round-robin fashion in each iteration, and the model with the ASPIN specific to the batch's anatomy is optimized. This training strategy allows each dataset to contribute to the anatomy-invariant representation learning and anatomy-specific representation learning. The pre-trained universal model is further fine-tuned in Step (S3) with combined $L_{MAE}$  and $ L_{AT}$ as its loss function: $L=L_{MAE}+  \omega L_{AT}$, where $\omega$ denotes a balancing coefficient between two terms. In Step (S4), when the new anatomy is available, the newly inserted ASPIN is trained with $L_{MAE}$ while freezing all the other parameters. 

\section{Experimental Results}
\subsection{Datasets And Network Configuration}
We used five public datasets to evaluate the proposed method. The brain and knee datasets are large-scale DICOM images from the NYU fastMRI \footnote{fastmri.med.nyu.edu} database \cite{knoll2020fastmri,zbontar2018fastmri}. These datasets are used to train the individual base models in Step (S1) and the universal model in Step (2-3). Three small-scale datasets are used to evaluate the generalization ability on new anatomies in Step (S4). They are cardiac dataset from ACDC challenge \cite{bernard2018deep}, abdomen dataset from CHAOS challenge \cite{kavur2021chaos,kavur2020comparison,CHAOSdata2019} and prostate dataset from Medical Segmentation Decathlon challenge \cite{simpson2019large}. In total we have 310,798 brain slices, 274,079 knee slices, 25,251 cardiac slices, 1,268 abdomen slices and 904 prostate slices. For each dataset, 80\% of the volumes are randomly extracted for training, 10\% for validation, and 10\% for testing.

In this study, all the data are emulated single-coil magnitude-only images. We cropped (brain and knee datasets) or resized (cardiac, abdomen and prostate datasets) the images into $320\times320$. Then, we retrospectively undersampled the Cartesian k-space with 1D Gaussian sampling pattern. The evaluation is conducted under acceleration rates of 4$\times$ and 6$\times$ using structural similarity index (SSIM) and peak signal-to-noise ratio (PSNR). We calculated and reported the mean value of the metrics for reconstructed test images.

All the experiments were carried out on NVIDIA SXM-2 Tesla V100 Accelerator Units, and the network was implemented on Pytorch. We trained networks using Adam optimizer with weight decay to prevent over-fitting. The size of all convolutional kernels is $3\times3$, the number of feature maps in CNN-t and uCNN-t are 32, and $\omega$ in the loss function of step 3 is empirically chosen to be 0.0001.  

\subsection{Algorithm Comparison}
We compared the proposed universal model with two cases. In the first case, we trained five D5C5 models (\say{Independent}) separately for each anatomy. This is the conventional approach that has good performance but uses the most parameters. In the second case, we mixed brain and knee datasets to train a single D5C5 model (\say{Shared}) and then tested its generalization ability to all datasets.

\begin{table}[ht]
\caption{Quantitative comparison under acceleration rates of $4\times$ and $6\times$.}\label{tab1}

\resizebox{\textwidth}{!}{%
\begin{tabular}{|l|l|rrrrrr|rrrrrr|}
\hline
\multicolumn{2}{|c|}{} & \multicolumn{6}{c|}{\textbf{PSNR (dB)}} &
\multicolumn{6}{c|}{\textbf{SSIM ($\%$)}} \\ \hline

\multicolumn{2}{|c|}{}&  \textbf{Brain} & \textbf{Knee} & \textbf{Card.} & \textbf{Abdo.} & \textbf{Pros.} & \textbf{Avg.} &  \textbf{Brain} & \textbf{Knee} & \textbf{Card.} & \textbf{Abdo.} & \textbf{Pros.} & \textbf{Avg.} \\  \hline

\multirow{5}{*}{\textbf{$4\times$}} & \textbf{Undersampled} & 31.60 & 30.45 &  29.20 &  33.52 &  26.28 & - &  78.69 & 76.47 & 80.62 &  75.58 & 71.55& - \\
 & \textbf{Independent} & 40.85 &  35.46 & 37.05  &  40.81   & 30.91  & 37.02 &  97.30 &  89.84& 96.07& 96.79& 86.03 & 93.21 \\
 & \textbf{Shared} & 40.78 & 35.43 & 35.18  & 41.15 & 30.30  & 36.57 & 97.30 & 89.87& 95.15 & 96.59& 84.86& 92.75 \\
 & \textbf{Proposed} & \textbf{43.16}  & \textbf{36.39} & \textbf{37.30}  & \textbf{42.76}  & \textbf{31.61} & \textbf{38.24}& \textbf{98.28}  &
  \textbf{91.39}& \textbf{96.51}& \textbf{98.20}& \textbf{87.72} & \textbf{94.42} \\

 & \textbf{w/o MD} & 42.80 & 36.25 & 37.10  & 42.59 & 31.51 & 38.05& 98.17 & 91.19& 96.42& 98.14& 87.51    & 94.29 \\\hline

\multirow{5}{*}{\textbf{$6\times$}} & \textbf{Undersampled} & 21.58 & 23.25 & 22.09 & 24.24& 21.55 & - & 58.47 & 60.90 & 61.87 & 51.66& 56.57 & - \\  & 
  \textbf{Independent} & 38.04 & 32.57 & 33.32 & 37.97 & 28.51& 34.08 & 95.48 & 85.01& 92.76  & 94.43 &79.33 & 89.40 \\  &
  \textbf{Shared} & 37.51 & 32.31 & 32.09 & 37.90 & 27.74 & 33.51 & 94.90 & 84.55 & 91.05 & 92.08 & 77.98 & 88.11 \\ &
  \textbf{Proposed} & \textbf{40.03} & \textbf{33.66} & \textbf{33.72}& \textbf{39.67}& \textbf{29.26} & \textbf{35.27} & \textbf{97.02}&
  \textbf{87.25} & \textbf{93.53} &
  \textbf{96.37} & \textbf{81.80} &
  \textbf{91.19} \\ 
 & \textbf{w/o MD} & 39.49 & 33.36 & 33.50 & 39.48 & 29.13 & 34.99 & 96.63 & 86.71 & 93.23 & 96.08 & 81.41 & 90.81 \\\hline
\end{tabular}%
}
\end{table}

The quantitative results are shown in Table \Ref{tab1}. Comparing along the columns, we observed the proposed method has the highest PSNR and SSIM on all datasets under both acceleration rates. The results of \say{Shared} is the lowest on all cases, which confirms that simply mixing all the datasets does not lead to an ideal model. The proposed method overcomes the issue and improved about 1.7 dB in PSNR on average under both acceleration rates. The improvement of the proposed method over \say{Shared} is also observed in SSIM, which records around 1.7$\%$ ($4\times$) and 3$\%$ ($6\times$) increase on average. The \say{Independent} model, optimised on each anatomy, achieved moderate performance on all datasets. The proposed method absorbed the knowledge from individual base models, and it outperformed \say{Independent} with considerable margin on brain and knee datasets. It is worth noting that, on the new cardiac, abdomen and prostate datasets, the proposed method also achieved superior performance to the \say{Independent} models. This indicates that the learned base of the universal model captures the anatomy-common knowledge and using ASPIN further allows adaptation to a new anatomy even with small-scale data. The effectiveness of the large-scale data is also seen in \cite{dar2020transfer,han2018deep,ouyang2019generalising}.

\begin{figure}
\includegraphics[width=\textwidth]{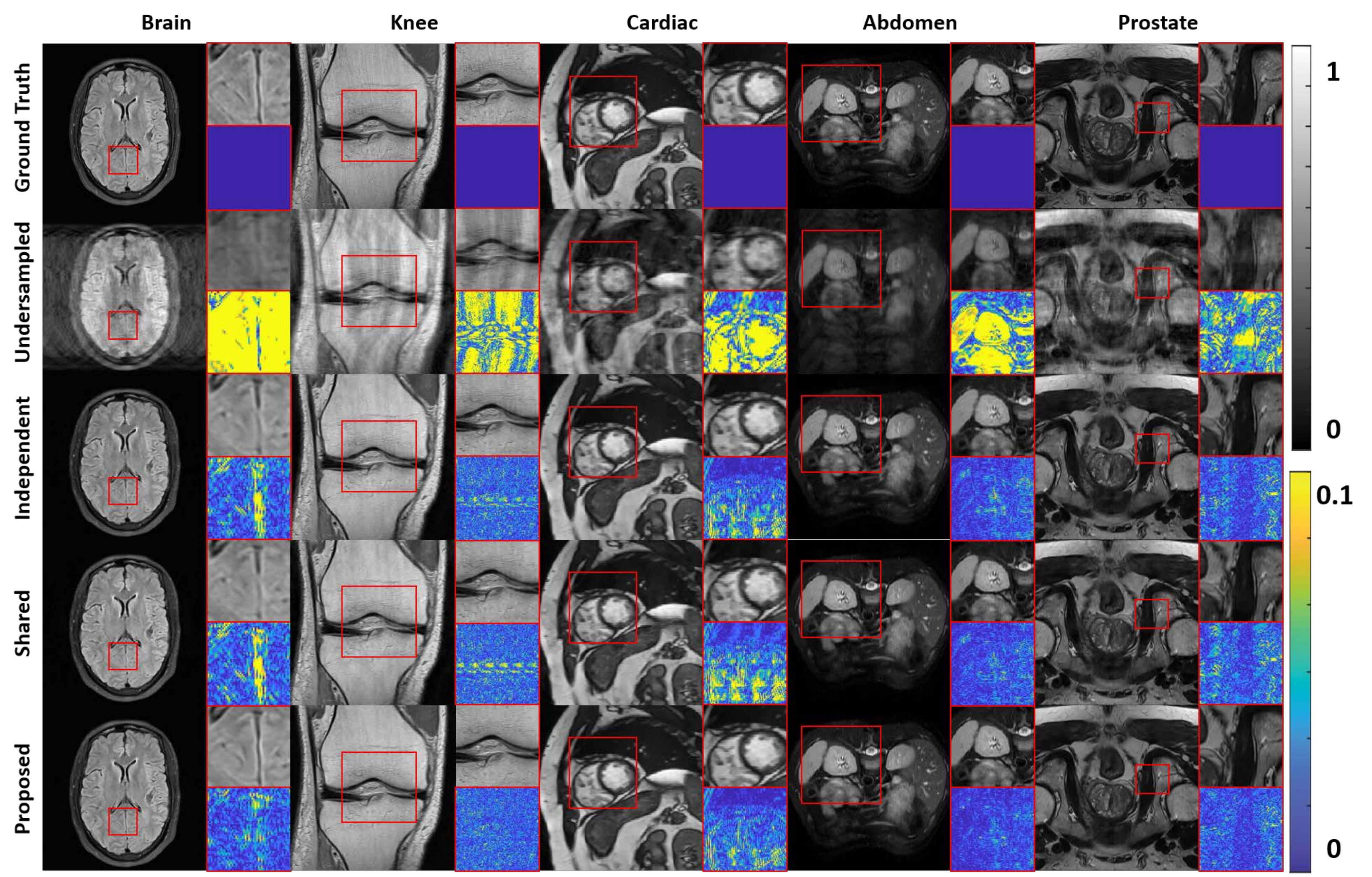}
\caption{Visualization of reconstructed images under acceleration rates of {$6\times$}. The zoomed-in area and the error maps are presented beside the images.} \label{fig2}
\end{figure}

We visualized the reconstructed images of different anatomies under $6\times$ acceleration in Fig.\Ref{fig2}. Both \say{Independent} and \say{Shared} reduced the artifacts presented in the undersampled images. \say{Proposed} further improved the reconstructed images, with clearer details and reduced error maps observed. 

\subsection{Ablation Study}
To examine the effectiveness of the proposed ASPIN and model distillation, we further conducted ablation study. We removed Step (S3) and trained a network only with Step (S2) for the multi-anatomy reconstruction, named as \say{w/o MD}. In other words, \say{w/o MD} improves Shared with ASPIN. When generalizing on the new anatomies, the newly inserted ASPIN in \say{w/o MD} still needs to be trained as in Step (S4).
 
The effectiveness of ASPIN is illustrated by comparing \say{Shared} with \say{w/o MD} in Table \Ref{tab1}. We observed that \say{w/o MD} improves around 2dB in PSNR on brain dataset and about 1dB on knee dataset, respectively. This shows the significance of ASPIN in compensating for statistical shift when the training dataset contains data of various anatomies. 
Comparing \say{Proposed} with \say{w/o MD}, we observed model distillation improved the quantitative results for brain and knee datasets on both acceleration rates. The improvement means incorporating the knowledge of individual base models in the universal model can further improve the performance. The increase in PSNR and SSIM is also consistently reported on cardiac, abdomen and prostate datasets, which further illustrates the importance of knowledge from individual base models. More ablation studies on different layer for distillation are provided in the supplementary materials, showing network with model distillation consistently outperformed \say{w/o MD}.

\subsection{Model Complexity}
A D5C5 model has 144,650 parameters, and we take it as an unit, or $1\times$. \say{Shared} uses one D5C5 model, so it has least of parameters. For \say{Independent}, reconstructing both brain and knee needs to train two D5C5 networks, leading to {$2\times$} parameters. To reconstruct a new anatomy, \say{Independent} needs another 144,650 parameters. In total, \say{Independent} needs {$5\times$} parameters for five anatomies. On the other hand, the proposed method only has additional parameters on ASPIN for each anatomy. One set of ASPIN for one anatomy has 1,280 parameters, so the proposed method only needs \textbf{less than $0.1\%$} of additional parameters for each anatomy. Overall, the proposed method has superior performance and only needs few additional parameters to reconstruct multiple anatomies.

\section{Conclusion}
We propose a novel universal deep neural network to reconstruct MRI of multiple anatomies. Experimental results demonstrate the proposed approach can reconstruct both brain and knee images, and it is easy to adapt to new small datasets while outperforming the individually trained models. While the current design is based on the popular D5C5 architecture, the proposed framework is extendable to other network architectures~\cite{zhou2020dudornet} for improved results. The current study is based on the single-coil magnitude-only images for proof-of-concept. In the future, the network can be adapted to multi-coil complex-valued datasets.

\bibliographystyle{splncs04}
\bibliography{arxiv_main}

\section{Supplementary Materials}
\renewcommand{\thetable}{S\arabic{table}}
\begin{table}[h]
\caption{Ablation studies on different layers in D5C5 for model distillation. The proposed method uses the third layer's knowledge for distillation. \textbf{w/o MD} is the results without model distillation. The proposed method with model distillation consistently outperformed \textbf{w/o MD}, with the third layer for distillation being the best.}\label{tab1 }% caption not exceed 100 words.

\resizebox{\textwidth}{!}{%
\begin{tabular}{|c|c|rrrrrr|rrrrrr|}
\hline

\multicolumn{2}{|c|}{} &\multicolumn{6}{c|}{\textbf{PSNR (dB)}} &
\multicolumn{6}{c|}{\textbf{SSIM ($\%$)}}\\
\hline

\textbf{Rates} & \textbf{Layer} & \textbf{Brain} & \textbf{Knee} & \textbf{Card.} & \textbf{Abdo.} & \textbf{Pros.} & \textbf{Avg.} &  \textbf{Brain} & \textbf{Knee} & \textbf{Card.} & \textbf{Abdo.} & \textbf{Pros.} & \textbf{Avg.} \\  \hline

\multirow{7}{*}{\textbf{$4\times$}}
 & \textbf{w/o MD} & 42.80 & 36.25 & 37.10  & 42.59 & 31.51 & 38.05& 98.17 & 91.19& 96.42& 98.14& 87.51    & 94.29 \\
& \textbf{\nth{1}} & 43.15 & 36.37 & 37.20 & 42.73 & 31.58& 38.21 &  98.27 & 91.37 & 96.48 & 98.19 & 87.67& 94.40 \\
 & \textbf{\nth{2}} & 43.14 &  36.37 & 37.23 & 42.74 & 31.59 &  38.21 & 98.26 &  91.37 & 96.49 & 98.20 & 87.68 & 94.40 \\
& \textbf{\nth{3}} & \textbf{43.16}  & \textbf{36.39} & \textbf{37.30}  & \textbf{42.76}  & \textbf{31.61} & \textbf{38.24}& \textbf{98.28}  &
  \textbf{91.39}& \textbf{96.51}& \textbf{98.20}& \textbf{87.72} & \textbf{94.42} \\
& \textbf{\nth{4}} & 43.14 & 36.38  & 37.19 & 42.74 & 31.60 & 38.21 &  98.27 & 91.37& 96.48 & 98.20 & 87.70 & 94.40 \\

& \textbf{\nth{5}}  & 43.14 & 36.38 & 37.24 & 42.75 & 31.60 & 38.22 & 98.27 & 91.38 & 96.49 & 98.20 & 87.70 & 94.41\\\hline

\multirow{6}{*}{\textbf{$6\times$}}
& \textbf{w/o MD} & 39.49 & 33.36 & 33.50 & 39.48 & 29.13 & 34.99 & 96.63 & 86.71 & 93.23 & 96.08 & 81.41 & 90.81 \\
& \textbf{\nth{1}} & 39.95 & 33.62 & 33.70 & 39.63 & 29.24 & 35.23 &  97.00 & 87.20 & 93.51 & 96.31 & 81.74 & 91.15\\
& \textbf{\nth{2}} & 39.94 & 33.61 & 33.69 & 39.62 & 29.25 & 35.22 &  96.99 & 87.22 & 93.50 & 96.28 & 81.76 & 91.15\\
& \textbf{\nth{3}} & \textbf{40.03} & \textbf{33.66} & \textbf{33.72}& \textbf{39.67}& \textbf{29.26} & \textbf{35.27} & \textbf{97.02}& \textbf{87.25} & \textbf{93.53} & \textbf{96.37} & \textbf{81.80} & \textbf{91.19} \\ 
& \textbf{\nth{4}} & 39.96 & 33.59 & 33.67 & 39.61 & 29.23 & 35.21 & 97.00 & 87.19 & 93.48 & 96.27 & 81.71 & 91.13\\

& \textbf{\nth{5}}  & 39.95 & 33.63  & 33.71 & 39.61 & 29.24 & 35.23 & 96.97 & 87.22 & 93.52 & 96.26 & 81.75 & 91.14 \\\hline
\end{tabular}}
\end{table}

\end{document}